\documentclass[11pt,a4paper]{article}
\usepackage[hyperref]{emnlp2019}

\usepackage{bbm}
\usepackage{booktabs}
\usepackage{color}
\usepackage{ulem}
\usepackage{graphicx}
\usepackage{latexsym}
\usepackage{svg}
\usepackage{textcomp}
\usepackage{times}
\usepackage{url}
\usepackage{tikz}
\usepackage{caption}
\usepackage[english]{babel}
\usepackage[autostyle, english = american]{csquotes}
\MakeOuterQuote{"}
\def\checkmark{\tikz\fill[scale=0.4](0,.35) -- (.25,0) -- (1,.7) -- (.25,.15) -- cycle;}
\usepackage{amsmath}
\usepackage{amssymb}
\usepackage{array,multirow}

\usepackage{titlesec}

\aclfinalcopy 

\DeclareMathOperator*{\argmax}{argmax}

\newcommand{\Asub}[1]{A(#1)}
\newcommand{\Esub}[1]{E(#1)}
\newcommand{\Lsub}[1]{L(#1)}
\newcommand{\Ssub}[1]{S(#1)}

\newcommand{\Agentsub}[1]{\textsc{Agent}(#1)}

\newcommand{\bertbasenospace}{\ensuremath{\text{BERT}_{\text{BASE}}}}
\newcommand{\bertbase}{\ensuremath{\text{BERT}_{\text{BASE}}}~}
\newcommand{\bertlargenospace}{\ensuremath{\text{BERT}_{\text{LARGE}}}}
\newcommand{\bertlarge}{\ensuremath{\text{BERT}_{\text{LARGE}}}~}

\newif\ifsubmit
\submitfalse
\ifsubmit
\newcommand{\cut}[1]{}
\newcommand{\dknote}[1]{}
\newcommand{\epnote}[1]{}
\newcommand{\kcnote}[1]{}
\newcommand{\jwnote}[1]{}
\newcommand{\rfnote}[1]{}
\newcommand{\sknote}[1]{}
\newcommand{\todo}[1]{}
\else
\newcommand{\cut}[1]{\sout{#1}}
\newcommand{\kcnote}[1]{\textcolor{purple}{\textbf{Cho: #1}}}
\newcommand{\dknote}[1]{\textcolor{pink}{\textbf{Douwe: #1}}}
\newcommand{\epnote}[1]{\textcolor{green}{\textbf{Ethan: #1}}}
\newcommand{\jwnote}[1]{\textcolor{orange}{\textbf{Jason: #1}}}
\newcommand{\rfnote}[1]{\textcolor{red}{\textbf{Rob: #1}}}
\newcommand{\sknote}[1]{\textcolor{brown}{\textbf{Sidd: #1}}}
\newcommand{\todo}[1]{\textcolor{blue}{\textbf{Todo: #1}}}
\fi

\title{Finding Generalizable Evidence by Learning to Convince Q\&A Models}

\author{Ethan Perez$^\dagger$ ~~ Siddharth Karamcheti$^\ddagger$\\\textbf{Rob Fergus}$^{\dagger\ddagger}$ ~~ \textbf{Jason Weston}$^{\dagger\ddagger}$ ~~ \textbf{Douwe Kiela}$^\ddagger$ ~~ \textbf{Kyunghyun Cho}$^{\dagger\ddagger\star}$\\
$^\dagger$New York University, $^\ddagger$Facebook AI Research, $^\star$CIFAR Azrieli Global Scholar\\
  {\tt perez@nyu.edu} \\}

\date{}

\begin{document}
\maketitle
\begin{abstract}
We propose a system that finds the strongest supporting evidence for a given answer to a question, using passage-based question-answering (QA) as a testbed.
We train evidence agents to select the passage sentences that most convince a pretrained QA model of a given answer, if the QA model received those sentences instead of the full passage.
Rather than finding evidence that convinces one model alone, we find that agents select evidence that generalizes; agent-chosen evidence increases the plausibility of the supported answer, as judged by other QA models and humans.
Given its general nature, this approach improves QA in a robust manner: using agent-selected evidence (i) humans can correctly answer questions with only $\sim$20\% of the full passage and (ii) QA models can generalize to longer passages and harder questions.
\end{abstract}

\section{Introduction}
\label{sec:Introduction}
\setcounter{section}{1}

\begin{figure}[t]
\centering
\includegraphics[width=\columnwidth]{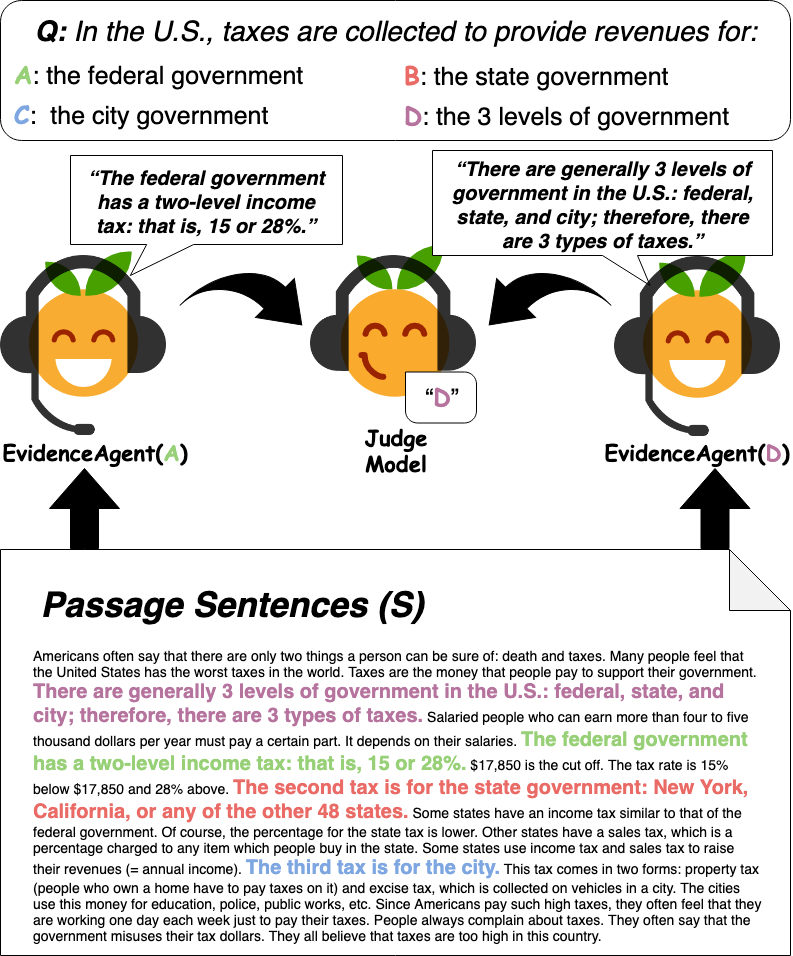}
\caption{
Evidence agents quote sentences from the passage to convince a question-answering judge model of an answer.
}
\label{fig:learning_to_persuade}
\end{figure}

\noindent There is great value in understanding the fundamental nature of a question~\cite{chalmers2015why}.
Distilling the core of an issue, however, is time-consuming. Finding the correct answer to a given question may require reading large volumes of text or understanding complex arguments.
Here, we examine if we can automatically discover the underlying properties of problems such as question answering by examining how machine learning models learn to solve that task.

We examine this question in the context of passage-based question-answering (QA).
Inspired by work in interpreting neural networks~\cite{lei2016rationalizing}, we have agents find a subset of the passage (i.e., supporting evidence) that maximizes a QA model's probability of a particular answer.
Each agent (one agent per answer) finds the sentences that a QA model regards as strong evidence for its answer, using either exhaustive search or learned prediction.
Figure~\ref{fig:learning_to_persuade} shows an example.

To examine to what extent evidence is general and independent of the model, we evaluate if humans and other models find selected evidence to be valid support for an answer too.
We find that, when provided with evidence selected by a given agent, both humans and models favor that agent's answer over other answers.
When human evaluators read an agent's selected evidence in lieu of the full passage, humans tend to select the agent-supported answer.

Given that this approach appears to capture some general, underlying properties of the problem, we examine if evidence agents can be used to assist human QA and to improve generalization of other QA models.
We find that humans can accurately answer questions on QA benchmarks, based on evidence for each possible answer, using only 20\% of the sentences in the full passage. 
We observe a similar trend with QA models: using only selected evidence, QA models trained on short passages can generalize more accurately to questions about longer passages, compared to when the models use the full passage.
Furthermore, QA models trained on middle-school reading comprehension questions generalize better to high-school exam questions by answering only based on the most convincing evidence instead of the full passage.
Overall, our results suggest that learning to select supporting evidence by having agents try to convince a judge model of their designated answer improves QA in a general and robust way.

\section{Learning to Convince Q\&A Models}
\label{sec:Learning to Convince Models}

Figure~\ref{fig:learning_to_persuade} shows an overview of the problem setup.
We aim to find the passage sentences that provide the most convincing evidence for each answer option, with respect to a given QA model (the \textit{judge}).
To do so, we are given a sequence of passage sentences $S=[\Ssub{1}, \dots, \Ssub{m}]$, a question $Q$, and a sequence of answer options $A=[\Asub{1}, \dots, \Asub{n}]$.
We train a judge model with parameters $\phi$ to predict the correct answer index $i^*$ by maximizing $p_{\phi}(\text{answer}=i^* | S, Q, A)$.

Next, we assign each answer $\Asub{i}$ to one evidence agent, $\Agentsub{i}$.
$\Agentsub{i}$ aims to find evidence $\Esub{i}$, a subsequence of passage sentences $S$ that the judge finds to support $\Asub{i}$.
For ease of notation, we use set notation to describe $\Esub{i}$ and $S$, though we emphasize these are ordered sequences.
$\Agentsub{i}$ aims to maximize the judge's probability on $\Asub{i}$ when conditioned on $\Esub{i}$ instead of $S$, i.e., $\argmax_{\Esub{i} \subseteq S} p_{\phi}(i| \Esub{i}, Q, A)$.
We now describe three different settings of having agents select evidence, which we use in different experimental sections (\S\ref{sec:Agents Select General Evidence}-\ref{sec:Evidence Agents Aid Human QA}).

\paragraph{Individual Sequential Decision-Making}

Since computing the optimal $\Esub{i}$ directly is intractable, a single $\Agentsub{i}$ can instead find a reasonable $\Esub{i}$ by making $T$ sequential, greedy choices about which sentence to add to $\Esub{i}$. In this setting, the agent ignores the actions of the other agents.
At time $t$, $\Agentsub{i}$ chooses index $e_{i,t}$ of the sentence in $S$ such that:
\begin{equation}
\label{eq:evidence agent objective}
    e_{i,t} = \argmax_{1 \leq e' \leq |S|} p_{\phi}(i|\{\Ssub{e'}\} \cup \Esub{i, t-1}, Q, A),
\end{equation}
where $\Esub{i,t}$ is the subsequence of sentences in $S$ that $\Agentsub{i}$ has chosen until time step $t$, i.e., \mbox{$\Esub{i,t} = \{\Ssub{e_{i,t}}\} \cup \Esub{i,t-1}$} with \mbox{$\Esub{i,0} = \varnothing$} and \mbox{$\Esub{i} = \Esub{i,T}$}.
It is a no-op to add a sentence $\Ssub{e_{i,t}}$ that is already in the selected evidence $\Esub{i,t-1}$.
The individual decision-making setting is useful for selecting evidence to support one particular answer.

\paragraph{Competing Agents: Free-for-All}
Alternatively, multiple evidence agents can compete at once to support unique answers, by each contributing part of the judge's total evidence.
Agent competition is useful as agents collectively select a pool of question-relevant evidence that may serve as a summary to answer the question.
Here, each of $\Agentsub{1}$, \ldots, $\Agentsub{n}$ finds evidence that would convince the judge to select its respective answer, $\Asub{1}$, \ldots, $\Asub{n}$.
$\Agentsub{i}$ chooses a sentence $\Ssub{e_{i,t}}$ by conditioning on all agents' prior choices:
\begin{equation*}
    e_{i,t} = \argmax_{1 \leq e' \leq |S|} p_{\phi}(i|\{\Ssub{e'}\} \cup E(*,t-1), Q, A),
\end{equation*}
where $E(*,t-1) = \cup_{j=1}^{n} \Esub{j,t-1}$.

Agents simultaneously select a sentence each, doing so sequentially for $t$ time steps, to jointly compose the final pool of evidence.
We allow an agent to select a sentence previously chosen by another agent, but we do not keep duplicates in the pool of evidence.
Conditioning on other agents' choices is a form of interaction that may enable competing agents to produce a more informative total pool of evidence.
More informative evidence may enable a judge to answer questions more accurately without the full passage.

\paragraph{Competing Agents: Round Robin}
Lastly, agents can compete round robin style, in which case we aggregate the outcomes of all $\binom{n}{2}$ pairs of answers $\{\Asub{i}, \Asub{j}\}$ competing.
Any given $\Agentsub{i}$ participates in $n-1$ rounds, each time contributing half of the sentences given to the judge.
In each one-on-one round, two agents select a sentence each at once.
They do so iteratively multiple times, as in the free-for-all setup.
To aggregate pairwise outcomes and compute an answer $i$'s probability, we average its probability over all rounds involving $\Agentsub{i}$:
\begin{equation*}
    \frac{1}{n-1}\sum_{j=1}^{n}{\mathbbm{1}{(i\neq j)} * p_{\phi}(i | \Esub{i} \cup \Esub{j}, Q, A)}
\end{equation*}

\subsection{Judge Models}
\label{ssec:Judge Models}
The judge model is trained on QA, and it is the model that the evidence agents need to convince. We aim to select diverse model classes, in order to: (i) test the generality of the evidence produced by learning to convince different models; and (ii) to have a broad suite of models to evaluate the agent-chosen evidence.
Each model class assigns every answer $\Asub{i}$ a score, where the predicted answer is the one with the highest score.
We use this score $\Lsub{i}$ as a softmax logit to produce answer probabilities.
Each model class computes $\Lsub{i}$ in a different manner.
In what follows, we describe the various judge models we examine.

\paragraph{TFIDF}
We define a function $\text{BoW}_{\text{TFIDF}}$ that embeds text into its corresponding TFIDF-weighted bag-of-words vector.
We compute the cosine similarity of the embeddings for two texts $\textbf{X}$ and $\textbf{Y}$:
\begin{equation*}
    \text{TFIDF}(\textbf{X}, \textbf{Y}) = \text{cos}(\text{BoW}_{\text{TFIDF}}(\textbf{X}), \text{BoW}_{\text{TFIDF}}(\textbf{Y}))
\end{equation*}

We define two model classes that select the answer most similar to the input passage sentences:
\mbox{$\Lsub{i}=\text{TFIDF}(S, [Q; \Asub{i}])$}, and 
\mbox{$\Lsub{i}=\text{TFIDF}(S, \Asub{i})$}.

\paragraph{\textit{fastText}}
We define a function $\text{BoW}_{\text{FT}}$ that computes the average bag-of-words representation of some text using \textit{fastText} embeddings~\cite{joulin2017bag}.
We use 300-dimensional \textit{fastText} word vectors pretrained on Common Crawl.
We compute the cosine similarity between the embeddings for two texts $\textbf{X}$ and $\textbf{Y}$ using:
\begin{equation*}
    \text{fastText}(\textbf{X}, \textbf{Y}) = \text{cos}(\text{BoW}_{\text{FT}}(\textbf{X}), \text{BoW}_{\text{FT}}(\textbf{Y}))
\end{equation*}

This method has proven to be a strong baseline for evaluating the similarity between two texts~\cite{perone2018evaluation}.
Using this function, we define a model class that selects the answer most similar to the input passage context:
\mbox{$\Lsub{i}=\text{fastText}(S, \Asub{i})$}.

\paragraph{BERT} $\Lsub{i}$ is computed using the multiple-choice adaptation of BERT~\cite{devlin2019bert,radford2018improving,si2019bert}, a pre-trained transformer network~\cite{vaswani2018attention}.
We fine-tune all BERT parameters during training.
This model predicts $\Lsub{i}$ using a trainable vector $v$ and BERT's first token embedding:
\mbox{$\Lsub{i} = v^\top \cdot \text{BERT}([S; Q; \Asub{i}])$}.

We experiment with both the~\bertbase model (12 layers) and \bertlarge (24 layers).
For training details, see Appendix~\ref{sec:Implementation Details}.

\subsection{Evidence Agents}
\label{ssec:Evidence Agents}

In this section, we describe the specific models we use as evidence agents.
The agents select sentences according to Equation~\ref{eq:evidence agent objective}, either exactly or via function approximation.

\paragraph{Search agent}
$\Agentsub{i}$ at time $t$ chooses the sentence $\Ssub{e_{i,t}}$ that maximizes $p_{\phi}(i|S(i,t),Q,A)$, after exhaustively trying each possible $\Ssub{e_{i,t}} \in S$.
Search agents that query TFIDF or fastText models maximize TFIDF or fastText scores directly (i.e., $\Lsub{i}$, rather than $p_{\phi}(i|S(i,t),Q,A)$).

\begin{table}[t]
\centering
\begin{tabular}{l | l l }
     \textit{\textbf{Predicting}} & \textbf{Loss} & \textbf{Target} \\
     \midrule
     Search & \textit{CE} & $S(e_{i,t})$ \\
     $p(i)$ & \textit{MSE} & $p_{\phi}(i | \{S(e')\} \cup \Esub{i,t}, Q, A)$ \\
     $\Delta p(i)$ & \textit{MSE} & $p_{\phi}(i | \{S(e')\} \cup \Esub{i,t}, Q, A)$ \\
     & & $\qquad~~- p_{\phi}(i | \Esub{i,t}, Q, A)$
\end{tabular}
\caption{
\label{tbl:learned-agents}
The loss functions and prediction targets for three learned agents. \textit{CE}: cross entropy. \textit{MSE}: mean squared error. $e'$ takes on integer values from 1 to $|S|$.
}
\end{table}

\paragraph{Learned agent}
We train a model to predict how a sentence would influence the judge's answer, instead of directly evaluating answer probabilities at test time.
This approach may be less prone to selecting sentences that exploit hard-to-predict quirks in the judge; humans may be less likely to find such sentences to be valid evidence for an answer (discussed in \S\ref{ssec:Human Evaluation of Evidence}).
We define several loss functions and prediction targets, shown in Table~\ref{tbl:learned-agents}.
Each forward pass, agents predict one scalar per passage sentence via end-of-sentence token positions.
We optimize these predictions using Adam~\cite{kingma2015adam} on one loss from Table~\ref{tbl:learned-agents}.
For $t > 1$, we find it effective to simply predict the judge model at $t=1$ and use this distribution for all time steps during inference.
This trick speeds up training by enabling us to precompute prediction targets using the judge model, instead of querying it constantly during training.

We use \bertbase for all learned agents.
Learned agents predict the \bertbase judge, as it is more efficient to compute than \bertlargenospace.
Each agent $\Agentsub{i}$ is assigned the answer $\Asub{i}$ that it should support.
We train one learned agent to find evidence for an arbitrary answer $i$.
We condition $\Agentsub{i}$ on $i$ using a binary indicator when predicting $L(i)$.
We add the indicator to BERT's first token segment indicator and embed it into vectors $\gamma$ and $\beta$; for each timestep's features $f$ from BERT, we scale and shift $f$ element-wise: $(\gamma*f)+\beta$~\citep{perez2018film,dumoulin2018feature-wise}.
See Appendix~\ref{sec:Implementation Details} for training details.

Notably, learning to convince a judge model does not require answer labels to a question.
Even if the judge only learns from a few labeled examples, evidence agents can learn to model the judge's behavior on more data and out-of-distribution data without labels.

\section{Experimental Setup}
\label{sec:Experimental Setup}

\subsection{Evaluating Evidence Agents}
\paragraph{Evaluation Desiderata}
An ideal evidence agent should be able to find evidence for its answer w.r.t. a judge, regardless (to some extent) of the specific answer it defends.
To appropriately evaluate evidence agents, we need to use questions with more than one defensible, passage-supported answer per question.
In this way, an agent's performance will not depend disproportionately on the answer it is to defend, rather than its ability to find evidence.

\paragraph{Multiple-choice QA: RACE and DREAM}
For our experiments, we use RACE~\cite{lai2017race} and DREAM~\cite{sun2018dream}, two multiple-choice, passage-based QA datasets.
Both consist of reading comprehension exams for Chinese students learning English; teachers explicitly designed answer options to be plausible (even if incorrect), in order to test language understanding.
Each question has 4 total answer options in RACE and 3 in DREAM.
Exactly one option is correct.
DREAM consists of 10K informal, dialogue-based passages.
RACE consists of 100K formal, written passages (i.e., news, fiction, or well-written articles).
RACE also divides into easier, middle school questions (29\%) and harder, high school questions (71\%).

\paragraph{Other datasets we considered}
Multiple-choice passage-based QA tasks are well-suited for our purposes.
Multiple-choice QA allows agents to support clear, dataset-curated possible answers.
In contrast,~\citet{sugawara2018what} show that 5-20\% of questions in extractive, span-based QA datasets have only one valid candidate option.
For example, some ``when'' questions are about passages with only one date.
\citeauthor{sugawara2018what} argue that multiple-choice datasets such as RACE do not have this issue, as answer candidates are manually created.
In preliminary experiments on SQuAD~\cite{rajpurkar2016squad}, we found that agents could only learn to convince the judge model when supporting the correct answer (one answer per question).

\subsection{Training and Evaluating Models}
\begin{table}[t]
  \centering
    \begin{tabular} {l|cc}
      \textit{\textbf{Judge Model}}             & \textbf{RACE}    & \textbf{DREAM} \\
      \midrule
      Random             & 25.0    & 33.3 \\
      \midrule
      TFIDF$(S, [Q; A])$ & 32.6    & 44.4 \\
      TFIDF$(S, A)$      & 31.6    & 44.5 \\
      fastText$(S, A)$   & 30.4    & 38.4 \\
      \bertbase          & 65.4    & 61.0 \\
      \bertlarge         & 69.4    & 64.9 \\
      Human Adult*       & 94.5    & 98.6 \\
    \end{tabular}
  \caption{
    RACE and DREAM test accuracy of various judge models using the full passage.
    Our agents use these models to find evidence.
    The models cover a spectrum of QA ability.
    (*) reports ceiling accuracy from original dataset papers.
  }
  \label{tab:judge_accuracies}
\end{table}

Our setup is not directly comparable to standard QA setups, as we aim to evaluate evidence rather than raw QA accuracy.
However, each judge model's accuracy is useful to know for analysis purposes.
Table~\ref{tab:judge_accuracies} shows model accuracies, which cover a broad range.
BERT models significantly outperform word-based baselines (TFIDF and fastText), and \bertlarge achieves the best overall accuracy.
No model achieves the estimated human ceiling for either RACE~\cite{lai2017race} or DREAM~\cite{sun2018dream}.

Our code is available at \url{https://github.com/ethanjperez/convince}.
We build off AllenNLP~\cite{gardner2017allennlp} using PyTorch~\cite{paszke2017automatic}.
For all human evaluations, we use Amazon Mechanical Turk via ParlAI~\cite{miller2017parlai}.
Appendix~\ref{sec:Implementation Details} describes preprocessing and training details.

\section{Agents Select General Evidence}
\label{sec:Agents Select General Evidence}
\begin{table*}[t]
  \centering
  \resizebox{.8\textwidth}{!}{
    \begin{tabular} {ll|ccc|ccc}
      & & \multicolumn{6}{c}{\textbf{\textit{How Often Human Selects Agent's Answer (\%)}}} \\
      & \multicolumn{1}{l}{} & \multicolumn{3}{|c}{\textbf{\textit{RACE}}} & \multicolumn{3}{|c}{\textbf{\textit{DREAM}}} \\
      & \multicolumn{1}{l}{\textbf{Evidence Sentence}} & \multicolumn{1}{|c}{} & \multicolumn{2}{c}{\textit{Agent Answer is}} & \multicolumn{1}{|c}{} & \multicolumn{2}{c}{\textit{Agent Answer is}} \\
      & \textbf{Selection Method}              & Overall       & Right         & Wrong         & Overall       & Right         & Wrong         \\
      \midrule
      \textbf{Baselines} & No Sentence Given       & 25.0          & 52.5          & 15.8          & 33.3          & 43.3          & 28.4          \\
      & Human Selection         & 41.6          & 75.1          & 30.4          & 50.7          & 84.9          & 33.5          \\
      \midrule
      \textbf{Search Agents} &
      TFIDF$(S,[Q;\Asub{i}])$ & 33.5          & 69.6          & 21.5          & 41.7          & 68.8          & 28.1          \\
      \hspace{.25cm} \textbf{querying...}& fastText$(S,\Asub{i})$  & 37.1          & 74.2          & 24.7          & 41.5          & 75.6          & 24.5          \\
      & TFIDF$(S,\Asub{i})$     & 38.0          & 71.4          & 26.9          & 43.4          & 75.2          & 27.6          \\
      & \bertbase               & 38.4          & 68.4          & 28.4          & 50.5          & 82.5          & 34.6          \\
      & \bertlarge              & 40.1          & 71.0          & 29.9          & 52.3          & 79.4          & 38.7          \\
      \textbf{Learned Agents} & Search                  & 40.0          & 71.0          & 29.7          & 49.1          & 78.3          & 34.6          \\
      \hspace{.25cm} \textbf{predicting...} & $p(i)$                  & 42.0          & 74.6          & 31.1          & 50.0          & 77.3          & 36.3          \\
      & $\Delta p(i)$           & 41.1          & 73.2          & 30.4          & 48.2          & 76.5          & 34.0          \\
    \end{tabular}
  }
  \caption{
    \textit{Human evaluation}:
    \textbf{Search Agents} select evidence by querying the specified judge model, and \textbf{Learned Agents} predict the strongest evidence w.r.t. a judge model (\bertbasenospace); humans then answer the question using the selected evidence sentence (without the full passage).
    Most agents do on average find evidence for their answer, right or wrong.
    Agents are more effective at supporting right answers.
  }
  \label{tab:human_on_quote}
\end{table*}

\subsection{Human Evaluation of Evidence}
\label{ssec:Human Evaluation of Evidence}
Would evidence that convinces a model also be valid evidence to humans?
On one hand, there is ample work suggesting that neural networks can learn similar patterns as humans do.
Convolutional networks trained on ImageNet share similarities with the human visual cortex~\cite{cadieu2014deep}.
In machine translation, attention learns to align foreign words with their native counterparts~\cite{bahdanau2015neural}.
On the other hand, neural networks often do not behave as humans do.
Neural networks are susceptible to adversarial examples---changes to the input which do or do not change the network's prediction in surprising ways~\cite{szegedy2014intriguing,jia2017adversarial,ribeiro2018semantically,alzantot2018generating}.
Convolutional networks rely heavily on texture~\cite{geirhos2018imagenettrained}, while humans rely on shape~\cite{landau1998importance}.
Neural networks trained to recognize textual entailment can rely heavily on dataset biases~\cite{gururangan2018annotation}.

\paragraph{Human evaluation setup}
We use human evaluation to assess how effectively agents select sentences that also make humans more likely to provide a given answer, when humans act as the judge.
Humans answer based only on the question $Q$, answer options $A$, and a single passage sentence chosen by the agent as evidence for its answer option $\Asub{i}$ (i.e., using the ``Individual Sequential Decision-Making'' scheme from \S\ref{sec:Learning to Convince Models}).
Appendix~\ref{sec:Human Evaluation Details} shows the interface and instructions used to collect evaluations.
For each of RACE and DREAM, we use 100 test questions and collect 5 human answers for each $(Q,\Asub{i})$ pair for each agent.
We also evaluate a human baseline for this task, where 3 annotators select the strongest supporting passage sentence for each $(Q,\Asub{i})$ pair.
We report the average results across 3 annotators.

\paragraph{Humans favor answers supported by evidence agents}
when shown that agent's selected evidence, as shown in Table~\ref{tab:human_on_quote}.\footnote{Appendix~\ref{sec:Human Evaluation of Agent Evidence by Question Category} shows results by question type.}
Without receiving any passage sentences, humans are at random chance at selecting the agent's answer (25\% on RACE, 33\% on DREAM), since agents are assigned an arbitrary answer.
For all evidence agents, humans favor agent-supported answers more often than the baseline (33.5-42.0\% on RACE and 41.7-50.5\% on DREAM).
For our best agents, the relative margin over the baseline is substantial.
In fact, these agents select evidence that is comparable to human-selected evidence.
For example, on RACE, humans select the target answer 41.6\% when provided with human-selected evidence, compared to 42\% evidence selected by the learned agent that predicts $p(i)$.

All agents support right answers more easily than wrong answers.
On RACE, the learned agent that predicts $p(i)$ finds strong evidence more than twice as often for correct answers than for incorrect ones (74.6\% vs. 31.1\%).
On RACE and DREAM both, BERT-based agents (search or learned agents) find stronger evidence than word-based agents do.
Humans tend to find that BERT-based agents select valid evidence for an answer, right or wrong.
On DREAM, word-based agents generally fail to find evidence for wrong answers compared to the no-sentence baseline (28.4\% vs. 24.5\% for a search-based fastText agent).

On RACE, learned agents that predict the \bertbase judge outperform search agents that directly query the \bertbase judge.
This effect may occur if search agents find an adversarial sentence that unduly affects the judge's answer but that humans do not find to be valid evidence.
Appendix~\ref{sec:Additional Evidence Agent Examples} shows one such example.
Learned agents may have difficulty predicting such sentences, without directly querying the judge.
Appendix~\ref{sec:Analysis} provides some analysis on why learned agents may find more general evidence than search agents do.
Learned agents are most accurate at predicting evidence sentences when the sentences have a large impact on the judge model's confidence in the target answer, and such sentences in turn are more likely to be found as strong evidence by humans.
On DREAM, search agents and learned agents perform similarly, likely because DREAM has 14x less training data than RACE.

\subsection{Model Evaluation of Evidence}
\label{ssec:Model Evaluation of Evidence}
\begin{figure}
    \centering
    \includegraphics[width=\columnwidth]{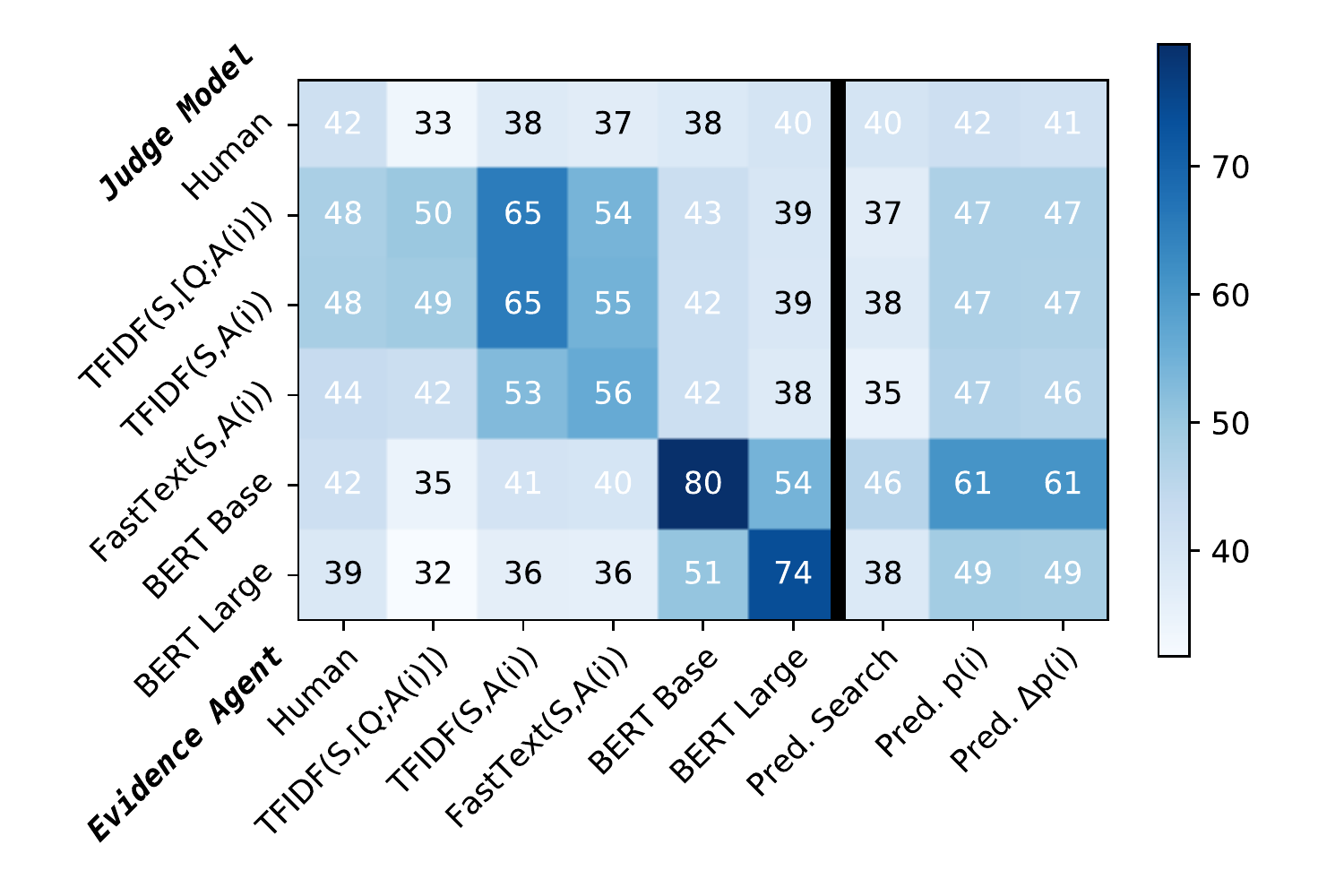}
    \caption{
    On RACE, how often each judge selects an agent's answer when given a single agent-chosen sentence.
    The black line divides learned agents (right) and search agents (left), with human evidence selection in the leftmost column.
    All agents find evidence that convinces judge models more often than a no-evidence baseline (25\%).
    Learned agents predicting $p(i)$ or $\Delta p(i)$ find the most broadly convincing evidence.
    }
    \label{fig:persuading_models}
\end{figure}

\paragraph{Evaluating an agent's evidence across models}
Beyond human evaluation, we test how general agent-selected evidence is, by testing this evidence against various judge models.
We expect evidence agents to most frequently convince the model they are optimized to convince, by nature of their direct training or search objective.
The more similar models are, the more we expect evidence from one model to be evidence to another.
To some extent, we expect different models to rely on similar patterns to answer questions.
Thus, evidence agents should sometimes select evidence that transfers to any model.
However, we would not expect agent evidence to transfer to other models if models only exploit method-specific patterns.

\paragraph{Experimental setup}
Each agent selects one evidence sentence for each $(Q,\Asub{i})$ pair.
We test how often the judge selects an agent's answer, when given this sentence, $Q$, and $A$.
We evaluate on all $(Q,\Asub{i})$ pairs in RACE's test set.
Human evaluations are on a 100 question subset of test.

\paragraph{Results}
Figure~\ref{fig:persuading_models} plots how often each judge selects an agent's answer.
Without any evidence, judge models are at random at choosing an agent's assigned answer (25\%).
All agents find evidence that convinces judge models more often than the no-evidence baseline.
Learned agents that predict $p(i)$ or $\Delta p(i)$ find the evidence most broadly considered convincing; other judge models select these agents' supported answers over 46\% of the time.
These findings support that evidence agents find general structure despite aiming to convince specific methods with their 
distinct properties.

Notably, evidence agents are not uniformly convincing across judge models.
All evidence agents are most convincing to the judge model they aim to convince; across any given agent's row, an agent's target judge model is the model which most frequently selects the agent's answer.
Search agents are particularly effective at finding convincing evidence w.r.t. their target judge model, given that they directly query this model.
More broadly, similar models find similar evidence convincing.
We find similar results for DREAM (Appendix~\ref{sec:Model Evaluation of Evidence on DREAM}).

\section{Evidence Agents Aid Generalization}
\label{sec:Evidence Agents Aid Generalization}

We have shown that agents capture method-agnostic evidence representative of answering a question (the strongest evidence for various answers).
We hypothesize that QA models can generalize better out of distribution to more challenging questions by exploiting evidence agents' capability to understand the problem.

Throughout this section, using various train/test splits of RACE, we train a \bertbase judge on easier examples (involving shorter passages or middle-school exams) and test its generalization to harder examples (involving longer passages or high-school exams).
Judge training follows \S\ref{ssec:Judge Models}.
We compare QA accuracy when the judge answers using (i) the full passage and (ii) only evidence sentences chosen by competing evidence agents.
We report results using the round robin competing agent setup described in \S\ref{sec:Learning to Convince Models}, as it resulted in higher generalization accuracy than free-for-all competition in preliminary experiments.
Each competing agent selects sentences up to a fixed, maximum turn limit; we experiment with 3-6 turns per agent (6-12 total sentences for the judge), and we report the best result.
We train learned agents (as described in \S\ref{ssec:Evidence Agents}) on the full RACE dataset without labels, so these agents can model the judge using more data and on out-of-distribution data.

For reference, we evaluate judge accuracy on a subsequence of randomly sampled sentences; we vary the number of sentences sampled from 6-12 and report the best result.
As a lower bound, we train an answer-only model to evaluate how effectively the QA model is using the passage sentences it is given.
As an upper bound, we evaluate our \bertbase judge trained on all of RACE, requiring no out-of-distribution generalization.

\subsection{Generalizing to Longer Passages}
\begin{table}[t]
  \centering
  \resizebox{\columnwidth}{!}{
    \begin{tabular} {ll|cc||c}
      &  & \multicolumn{2}{c}{\textbf{\textit{RACE}}} & \textbf{$\rightarrow$\textit{DREAM}} \\
      \textbf{Train} &   & \multicolumn{3}{c}{\textit{Sentences in Passage}} \\
      \textbf{Data} & Sentence Selection  & $\leq 12$        & $\geq 27$        & $\geq 27$      \\
      \midrule
        \textbf{All} & Full Passage  & 64.7          & 60.0          & 71.2        \\
      \midrule
        \textbf{RACE}
        & None (Answer-only)           & 36.1          & 40.2          & 38.5        \\
        $|S|\leq 12$ & Full Passage of Subset & 57.4   & 44.1          & 65.0        \\\vspace{.2cm}
        & Random Sentences             & 49.2          & 44.7          & 48.2        \\
        & TFIDF$(S,[Q;\Asub{i}])$      & 57.2          & 48.0          & 67.3        \\
        & fastText$(S,\Asub{i})$       & \textbf{57.7} & \textbf{50.2}          & 64.2        \\
        & TFIDF$(S,\Asub{i})$          & 57.1          & 47.9          & 64.6        \\
        & Search over \bertbase        & 56.7          & 49.6          & \textbf{68.9} \\
        & Predict \bertbase $p(i)$  & 56.7          & 50.0 & 66.9        \\
    \end{tabular}
  }
  \caption{
    We train a judge on short RACE passages and test its generalization to long passages.
    The judge is more accurate on long passages when it answers based on only sentences chosen by competing agents (last 5 rows) instead of the full passage.
    BERT-based agents aid generalization even under test-time domain shift (from RACE to DREAM).
  }
  \label{tab:generalizing_to_longer_passages}
\end{table}

We train a judge on RACE passages averaging 10 sentences long (all training passages each with $\leq$12 sentences); this data is roughly $\frac{1}{10}$th of RACE.
We test the judge on RACE passages averaging 30 sentences long.

\paragraph{Results}
Table~\ref{tab:generalizing_to_longer_passages} shows the results.
Using the full passage, the judge outperforms an answer-only BERT baseline by 4\% (44.1\% vs. 40.2\%).
When answering using the smaller set of agent-chosen sentences, the judge outperforms the baseline by 10\% (50.2\% vs. 40.2\%), more than doubling its relative use of the passage.
Both search and learned agents aid the judge model in generalizing to longer passages.
The improved generalization is not simply a result of the judge using a shorter passage, as shown by the random sentence selection baseline (44.7\%).

\subsection{Generalizing Across Domains}

We examine if evidence agents aid generalization even in the face of domain shift.
We test the judge trained on short RACE passages on long passages from \textit{DREAM}.
We use the same evidence agents from the previous subsection; the learned agent is trained on RACE only, and we do not fine-tune it on DREAM to test its generalization to finding evidence in a new domain.
DREAM passages consist entirely of dialogues, use more informal language and shorter sentences, and emphasize general world knowledge and commonsense reasoning~\cite{sun2018dream}.
RACE passages are more formal, written articles (e.g. news or fiction).

\paragraph{Results}
Table~\ref{tab:generalizing_to_longer_passages} shows that BERT-based evidence agents aid generalization even under domain shift.
The model shows notable improvements for RACE $\rightarrow$ DREAM transfer when it predicts from BERT-based agent evidence rather than the full passage (65.0\% vs. 68.9\%). These results support that our best evidence agents capture something fundamental to the problem of QA, despite changes in e.g. content and writing style.

\subsection{Generalizing to Harder Questions}
\begin{table}[t]
\small
  \centering
    \begin{tabular} {ll|cc}
      \textbf{Train} &  & \multicolumn{2}{|c}{\textit{School Level}} \\
      \multicolumn{1}{l}{\textbf{Data}}{} & Sentence Selection & Middle        & High          \\
      \midrule
        \textbf{All} & Full Passage & 70.8          & 63.2          \\
      \midrule
        \textbf{Middle} & None (Answer-only)            & 38.9          & 40.2          \\
        \textbf{School} & Full Passage of Subset        & 66.2          & 50.7          \\\vspace{.15cm}
        \textbf{only} & Random Sentences    & 54.8          & 47.0          \\
        & TFIDF$(S,[Q;\Asub{i}])$      & 65.1          & 50.4          \\
        & fastText$(S,\Asub{i})$       & 64.6          & 50.8          \\
        & TFIDF$(S,\Asub{i})$          & 64.9          & 51.0          \\
        & Search over \bertbase     & 67.0 & \textbf{53.0} \\
        & Predict \bertbase $p(i)$ & \textbf{67.3}          & 51.9          \\ 
    \end{tabular}
  \caption{
    \textit{Generalizing to harder questions}:
    We train a judge to answer questions with RACE's Middle School exam questions only.
    We test its generalization to High School exam questions.
    The judge is more accurate when using evidence agent sentences (last 5 rows) rather than the full passage.
  }
  \label{tab:generalizing_to_harder_questions}
\end{table}
\begin{figure}
    \centering
    \includegraphics[width=\columnwidth]{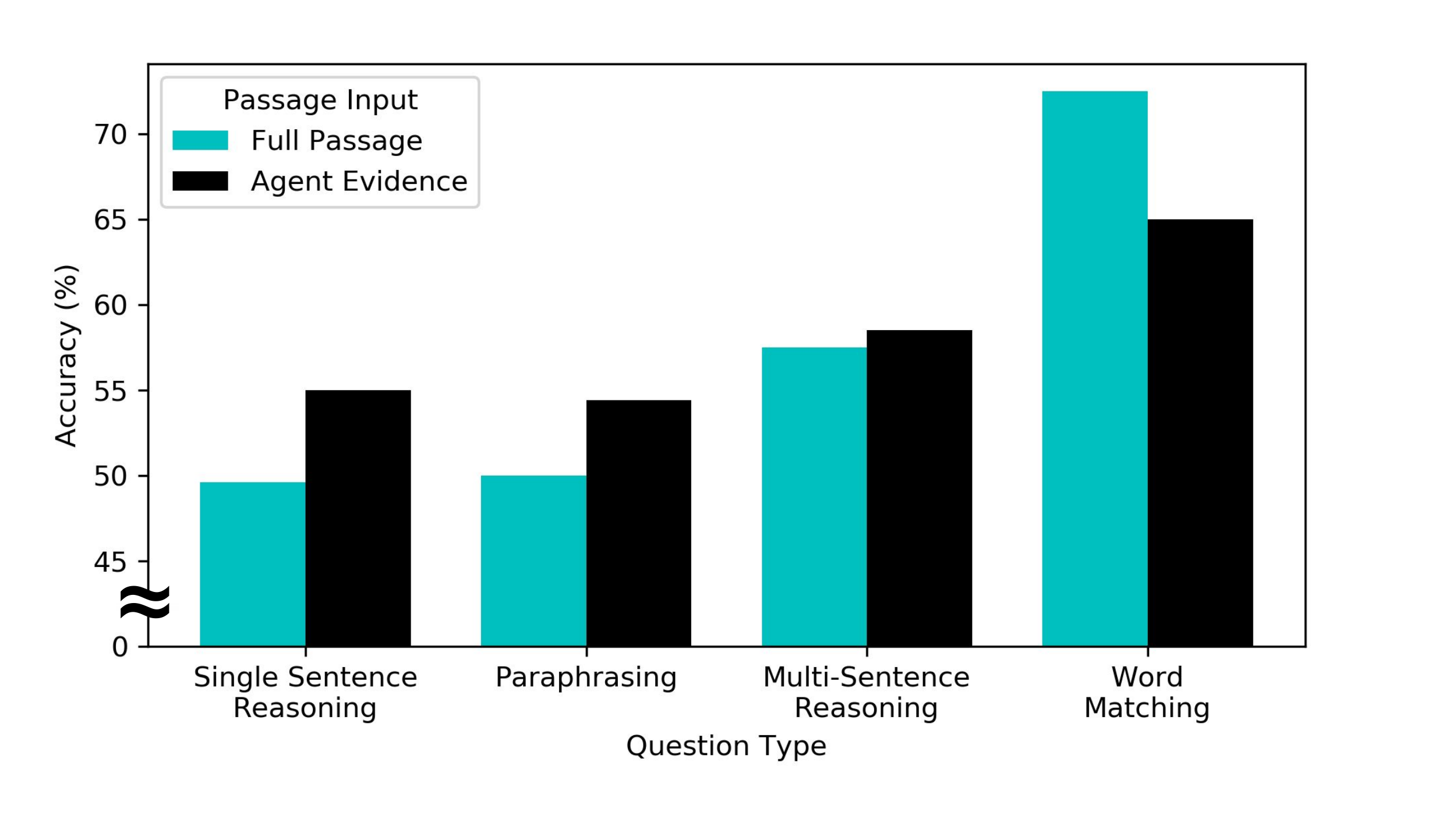}
    \caption{
        \textit{Generalizing to harder questions by question type}:
        We train a judge on RACE Middle School questions and test its generalization to RACE High School questions.
        To predict the answer, the judge uses either the full passage or evidence sentences chosen by a BERT-based search agent.
        The worse the judge does on a question category using the full passage, the better it does when using the agent-chosen sentences.
    }
    \label{fig:generalizing_to_harder_questions_by_question_type}
\end{figure}

Using RACE, we train a judge on middle-school questions and test it on high-school questions.

\paragraph{Results} Table~\ref{tab:generalizing_to_harder_questions} shows that 
the judge generalizes to harder questions better by using evidence from either search-based BERT agents (53.0\%) or learned BERT agents (51.9\%) compared to using the full passage directly (50.7\%) or to search-based TFIDF and fastText agents (50.4\%-51.0\%).
Figure~\ref{fig:generalizing_to_harder_questions_by_question_type} shows that the improved generalization comes from questions the model originally generalizes worse on.
Simplifying the passage by providing key sentences may aid generalization by e.g. removing extraneous or distracting sentences from passages with more uncommon words or complex sentence structure.
Such improvements come at the cost of accuracy on easier, word-matching questions, where it may be simpler to answer with the full passage as seen in training.

\section{Evidence Agents Aid Human QA}
\label{sec:Evidence Agents Aid Human QA}
As observed in \S\ref{ssec:Human Evaluation of Evidence}, evidence agents more easily support right answers than wrong ones.
Furthermore, evidence agents do aid QA models in generalizing systematically when all answer evidence sentences are presented at once.
We hypothesize that when we combine all evidence sentences, humans prefer to choose the correct answer.

\paragraph{Human evaluation setup}

Evidence agents compete in a free-for-all setup (\S\ref{sec:Learning to Convince Models}), and the human acts as the judge.
We evaluate how accurately humans can answer questions based only on agent sentences.
Appendix~\ref{sec:Human Evaluation Details} shows the annotation interface and instructions.
We collect 5 human answers for each of the 100 test questions.

\paragraph{Humans can answer using evidence sentences alone}

Shown in Table~\ref{tab:human_on_quotes},
humans correctly answer questions using many fewer sentences (3.3 vs. 18.2 on RACE, 2.4 vs. 12.2 on DREAM); they do so while maintaining 90\% of human QA accuracy on the full passage (73.2\% vs. 82.3\% on RACE, 83.8\% vs. 93.0\% on DREAM).
Evidence agents, however, vary in how effectively they aid human QA, compared to answer-agnostic evidence selection.
On DREAM, humans answer with 79.1\% accuracy using the sentences most similar to the question alone (via fastText), while achieving lower accuracy when using the \bertlarge search agent's evidence (75.0\%) and higher accuracy when using the \bertbase search agent's evidence (83.8\%).
We explain the discrepancy by examining how effective agents are at supporting right vs. wrong answers (Table~\ref{tab:human_on_quote} from \S\ref{ssec:Human Evaluation of Evidence}); 
\bertbase is more effective than \bertlarge at finding evidence for right answers (82.5\% vs. 79.4\%) and less effective at finding evidence for wrong answers (34.6\% vs. 38.7\%).

\begin{table}[t]
  \centering
  \resizebox{\columnwidth}{!}{
    \begin{tabular} {ll|cc}
      \textbf{Sentences Shown}& & \multicolumn{2}{c}{\textbf{\textit{Human Acc. (\%)}}} \\
      \hspace{.25cm} \textit{Selection Type} & Selection Method & \textbf{RACE} & \textbf{DREAM} \\
      \midrule
      \textbf{Full Passage} & Full Passage                    & 82.3          & 93.0          \\
      \textbf{No Passage} & Answer-only           & 52.5          & 43.3          \\
      \midrule
      \textbf{Subset (\texttildelow 20\%)} & Human Selection  & \textbf{73.5} & 82.3          \\  
      \hspace{.25cm} \textit{Answer-Free} & First $n$ Sentences        & 61.8          & 68.5          \\
      \hspace{.25cm} \textit{Selection} & TFIDF$(S, Q)$              & 69.2          & 77.5          \\
      & fastText$(S, Q)$           & 69.7          & 79.1          \\
      \hspace{.25cm} \textit{Search Agent} & TFIDF$(S, [Q; \Asub{i}])$  & 66.1          & 70.0          \\
      \hspace{.25cm} \textit{Selection}& TFIDF$(S, \Asub{i})$       & 73.2          & 77.0          \\
      & fastText$(S, \Asub{i})$    & 73.2          & 77.3          \\
      & \bertbase                  & 69.9          & \textbf{83.8} \\
      & \bertlarge                 & 72.4          & 75.0          \\
      \hspace{.25cm} \textit{Learned Agent} & Predicting Search          & 66.5          & 80.0          \\
      \hspace{.25cm} \textit{Selection}& Predicting $p(i)$          & 71.6          & 77.8          \\
      & Predicting $\Delta p(i)$   & 65.7          & 81.5          \\
    \end{tabular}
  }
  \vspace{-2mm}
  \caption{
    \textit{Human accuracy using evidence agent sentences}:
    Each agent selects a sentence supporting its own answer.
    Humans answer the question given these agent-selected passage sentences only.
    Humans still answer most questions correctly, while reading many fewer passage sentences.
  }
  \label{tab:human_on_quotes}
  
  \vspace{-4mm}
\end{table}
\section{Related Work}
\label{sec:Related Work}

Here, we discuss further related work, beyond that discussed in \S\ref{ssec:Human Evaluation of Evidence} on (dis)similarities between patterns learned by humans and neural networks.

\paragraph{Evidence Extraction}
Various papers have explored the related problem of extracting evidence or summaries to aid downstream QA.
\citet{wang2018evidence-extraction} concurrently introduced a neural model that extracts evidence specifically for the correct answer, as an intermediate step in a QA pipeline.
Prior work uses similar methods to explain what a specific model has learned~\cite{lei2016rationalizing,li2016understanding,yu2019learning}.
Others extract evidence to improve downstream QA efficiency over large amounts of text~\cite{choi2017coarse,wang2018r3,wang2018evidence-aggregation}.
More broadly, extracting evidence can facilitate fact verification~\cite{thorne2018fever} and debate.\footnote{IBM Project Debater: \url{www.research.ibm.com/artificial-intelligence/project-debater}}

\paragraph{Generic Summarization}
In contrast, various papers focus primarily on summarization rather than QA, using downstream QA accuracy only as a reward to optimize generic (question-agnostic) summarization models~\cite{arumae2018reinforced,arumae2019guiding,eyal2019question}.

\paragraph{Debate}
Evidence extraction can be viewed as a form of debate, in which multiple agents support different stances~\citep{irving2018ai,irving2019ai}.
\citet{chen2018cicero} show that evidence-based debate improves the accuracy of crowdsourced labels, similar to our work which shows its utility in natural language QA.

\section{Conclusion}
\label{sec:Conclusion}
We examined if it was possible to automatically distill general insights for passage-based question answering, by training evidence agents to convince a judge model of any given answer. Humans correctly answer questions while reading only 20\% of the sentences in the full passage, showing the potential of our approach for assisting humans in question answering tasks. We examine how selected evidence affects the answers of humans as well as other QA models, and we find that agent-selected evidence is generalizable. We exploit these capabilities by employing evidence agents to facilitate QA models in generalizing to longer passages and out-of-distribution test sets of qualitatively harder questions.

\section*{Acknowledgments}
EP was supported by the NSF Graduate Research Fellowship and ONR grant N00014-16-1-2698.
KC thanks support from eBay and NVIDIA.
We thank Adam Gleave, David Krueger, Geoffrey Irving, Katharina Kann, Nikita Nangia, and Sam Bowman for helpful conversations and feedback.
We thank Jack Urbanek, Jason Lee, Ilia Kulikov, Ivanka Perez, Ivy Perez, and our Mechanical Turk workers for help with human evaluations.

\bibliography{emnlp2019}
\bibliographystyle{acl_natbib}

\clearpage
\appendix

\begin{table}[th!]
    \centering
    \footnotesize
    \begin{tabular}{p{0.2cm}p{6.5cm}}
        \toprule
        \textbf{Passage~(DREAM)} & \\
        \midrule
        \textbf{W}: & \hspace{.05cm} What changes do you think will take place in the next 50 years? \\
        \textbf{M}: & \hspace{.05cm} I imagine that the greatest change will be the difference between humans and machines. \\
        \textbf{W}: & \hspace{.05cm} What do you mean? \\
        \textbf{M}: & \hspace{.05cm} I mean it will be harder to tell the difference between the human and the machine. \\
        \textbf{W}: & \hspace{.05cm} Can you describe it more clearly? \\
        \textbf{M}: & \hspace{.05cm} As science develops, it will be possible for all parts of one's body to be replaced. \textbf{\textcolor{magenta}{A computer will work like the human brain.}} The computer can recognize one's feelings, and act in a feeling way. \\
        \textbf{W}: & \hspace{.05cm} You mean man-made human beings will be produced? Come on! That's out of the question! \\
        \textbf{M}: & \hspace{.05cm} Don't get excited, please. \textbf{\textcolor{cyan}{That's only my personal imagination!}} \\
        \textbf{W}: & \hspace{.05cm} Go on, please. I won't take it seriously. \\
        \textbf{M}: & \hspace{.05cm} We will then be able to create a machine that is a copy of ourselves. We'll appear to be alive long after we are dead. \\
        \textbf{W}: & \hspace{.05cm} What a ridiculous idea! \\
        \textbf{\textcolor{orange}{\textbf{M}:}} & \hspace{.05cm} \textbf{\textcolor{orange}{It's possible that a way will be found to put our spirit into a new body.}} Then, we can choose to live as long as we want. \\
        \textbf{W}: & \hspace{.05cm} In that case, the world would be a hopeless mess! \\
        \midrule
        \textit{\textbf{Q}:} & \textit{\textbf{What are the two speakers talking about?}}\\
        \textcolor{magenta}{\textbf{A.}} & \textcolor{magenta}{\textbf{Computers in the future.}}\\
        \textcolor{cyan}{\textbf{B.}} & \textcolor{cyan}{\textbf{People's imagination.}}\\
        \textcolor{orange}{\textbf{C.}} & \textcolor{orange}{\textbf{Possible changes in the future.}}\checkmark\\
        \bottomrule
    \end{tabular}
    \caption{
    An example from our best evidence agent on DREAM, a search agent using \bertlargenospace.
    Each evidence agent has chosen a sentence (in color) that convinces a \bertlarge judge model to predict the agent's designated answer with over 99\% confidence.
    }
    \label{tab:examples/dream/human_eval_overview}
\end{table}
\begin{table*}[th!]
    \centering
    \footnotesize
    \begin{tabular}{p{\textwidth}}
        \toprule
        \textbf{Passage (RACE)} \\
        \midrule
        \tiny
        Who doesn't love sitting beside a cosy fire on a cold winter's night? Who doesn't love to watch flames curling up a chimney? Fire is one of man's greatest friends, but also one of his greatest enemies.
        \small \textcolor{cyan}{\textbf{Many big fires are caused by carelessness.}} \textcolor{magenta}{\textbf{A lighted cigarette thrown out of a car or train window or a broken bottle lying on dry grass can start a fire.}} \textcolor{orange}{\textbf{Sometimes, though, a fire can start on its own.}} \tiny
        Wet hay can begin burning by itself. This is how it happens: the hay starts to rot and begins to give off heat which is trapped inside it. Finally, it bursts into flames.
        \small \textcolor{brown}{\textbf{That's why farmers cut and store their hay when it's dry.}} \tiny
        Fires have destroyed whole cities. In the 17th century, a small fire which began in a baker's shop burnt down nearly every building in London. Moscow was set on fire during the war against Napoleon. This fire continued burning for seven days. And, of course, in 64 A.D. a fire burnt Rome. Even today, in spite of modern fire-fighting methods, fire causes millions of pounds' worth of damage each year both in our cities and in the countryside. It has been wisely said that fire is a good servant but a bad master.\\
        \midrule
        \textbf{Q}: Many big fires are caused...\\
        \textcolor{magenta}{\textbf{A.} by cigarette} \hspace{.2cm} \textcolor{orange}{\textbf{B.} by their own} \hspace{.2cm} \textcolor{brown}{\textbf{C.} by dry grass} \hspace{.2cm} \textcolor{cyan}{\textbf{D.} by people's carelessness} \checkmark\\
        \bottomrule
    \end{tabular}
    \caption{
    In this example, each answer's agent has chosen a sentence (in color) that individually influenced a neural QA model to answer in its favor.
    When human evaluators answer the question using only one agent's sentence, evaluators select the agent-supported answer.
    When humans read all 4 agent-chosen sentences together, they correctly answer ``D'', without reading the full passage.
    }
    \label{tab:examples/race/human_eval_overview_fire}
\end{table*}
\begin{table*}[th!]
    \centering
    \footnotesize
    \begin{tabular}{p{\textwidth}}
        \toprule
        \textbf{Passage (RACE)} \\
        \midrule
        \tiny
        Yueyang Tower lies in the west of Yueyang City, near the Dongting Lake. It was first built for soldiers to rest on and watch out. In the Three Kingdoms Period, Lu Su, General of Wu State, trained his soldiers here.
        \small \textcolor{magenta}{\textbf{In 716, Kaiyuan of Tang Dynasty, General Zhang Shuo was sent to defend at Yuezhou and he rebuilt it into a tower named South Tower, and then Yueyang Tower.}} \textcolor{cyan}{\textbf{In 1044, Song Dynasty, Teng Zijing was stationed at Baling Jun, the ancient name of Yueyang City.}} \tiny
        In the second year, he had the Yueyang Tower repaired and had poems by famous poets written on the walls of the tower. Fan Zhongyan, a great artist and poet, was invited to write the well - known poem about Yueyang Tower.
        \small \textcolor{brown}{\textbf{In his \textit{A Panegyric of the Yueyang Tower}, Fan writes: "Be the first to worry about the troubles across the land, the last to enjoy universal happiness."}} \tiny
        His words have been well - known for thousands of years and made the tower even better known than before. The style of Yueyang Tower is quite special. The main tower is 21.35 meters high with 3 stories, flying eave and wood construction, the helmet-roof of such a large size is a rarity among the ancient architectures in China.
        \small \textcolor{orange}{\textbf{Entering the tower, you'll see "Dongting is the water of the world, Yueyang is the tower of the world".}} \tiny
        Moving on, there is a platform that once used as the training ground for the navy of Three-Kingdom Period general Lu Su. To its south is the Huaifu Pavilion in honor of Du Fu. Stepping out of the Xiaoxiang Door, the Xianmei Pavilion and the Sanzui Pavilion can be seen standing on two sides. In the garden to the north of the tower is the tomb of Xiaoqiao, the wife of Zhou Yu.\\
        \midrule
        \textbf{Q}: Yueyang Tower was once named...\\
        \textcolor{magenta}{\textbf{A.} South Tower} \checkmark \hspace{.2cm}
        \textcolor{orange}{\textbf{B.} Xianmei Tower} \hspace{.2cm}
        \textcolor{brown}{\textbf{C.} Sanzui Tower} \hspace{.2cm}
        \textcolor{cyan}{\textbf{D.} Baling Tower}\\
        \bottomrule
    \end{tabular}
    \caption{
    An example where each answer's search agents successfully influences the answerer to predict that agent's answer; however, the supporting sentence for ``B'' and for ``C'' are not evidence for the corresponding answer.
    These search agents have found adversarial examples in the passage that unduly influence the answerer.
    Thus, it can help to present the answerer model with evidence for $2+$ answers at once, so the model can weigh potentially adversarial evidence against valid evidence.
    In this case, the model correctly answers ``B'' when predicting based on all 4 agent-chosen sentences.
    }
    \label{tab:examples/race/adversarial_example_persuades_judge}
\end{table*}
\begin{table*}[th!]
    \centering
    \footnotesize
    \begin{tabular}{p{\textwidth}}
        \toprule
        \textbf{Passage (RACE)} \\
        \midrule
        \tiny
        A desert is a beautiful land of silence and space.
        \small \textcolor{brown}{\textbf{The sun shines, the wind blows, and time and space seem endless.}} \tiny
        Nothing is soft. The sand and rocks are hard, and many of the plants even have hard needles instead of leaves. \small \textcolor{orange}{\textbf{The size and location of the world's deserts are always changing.}} \tiny
        Over millions of years, as climates change and mountains rise, new dry and wet areas develop. But within the last 100 yeas, deserts have been growing at a frightening speed. This is partly because of natural changes, but the greatest makers are humans.
        \small \textcolor{cyan}{\textbf{Humans can make deserts, but humans can also prevent their growth.}} \textcolor{magenta}{\textbf{Algeria Mauritania is planting a similar wall around Nouakchott, the capital.}} \tiny
        Iran puts a thin covering of petroleum on sandy areas and plants trees. The oil keeps the water and small trees in the land, and men on motorcycles keep the sheep and goats away. The USSR and India are building long canals to bring water to desert areas.\\
        \midrule
        \textbf{Q}: Which of the following is NOT true?\\
        \textcolor{magenta}{\textbf{A.} The greatest desert makers are humans.} \hspace{.2cm}
        \textcolor{orange}{\textbf{B.} There aren't any living things in the deserts.} \checkmark\\
        \textcolor{brown}{\textbf{C.} Deserts have been growing quickly.} \hspace{.2cm}
        \textcolor{cyan}{\textbf{D.} The size of the deserts is always changing.}\\
        \bottomrule
    \end{tabular}
    \caption{
    In this example, the answerer correctly predicts ``B,'' no matter the passage sentence (in color) a search agent provides.
    This behavior occurred in several cases where the question and answer options contained a strong bias in wording that cues the right answer.
    Statements including ``all,'' ``never,'' or ``there aren't any'' are often false, which in this example signals the right answer.
    \citet{gururangan2018annotation} find similar patterns in natural language inference data, where ``no,'' ``never,'' and ``nothing'' strongly signal that one statement contradicts another.
    }
    \label{tab:examples/race/cannot_persuade_judge}
\end{table*}

\section{Additional Evidence Agent Examples}
\label{sec:Additional Evidence Agent Examples}

We show additional examples of evidence agent sentence selections in Table~\ref{tab:examples/dream/human_eval_overview} (DREAM), as well as Tables~\ref{tab:examples/race/human_eval_overview_fire},~\ref{tab:examples/race/adversarial_example_persuades_judge}, and~\ref{tab:examples/race/cannot_persuade_judge} (RACE).

\section{Implementation Details}
\label{sec:Implementation Details}

\subsection{Preprocessing}
We use the BERT tokenizer to tokenize the text for all methods (including TFIDF and fastText).
To divide the passage into sentences, we use the following tokens as end-of-sentence markers: ``.'', ``?'', ``!'', and the last passage token.
For BERT, we use the required WordPiece subword tokenization~\cite{schuster2012japanese}.
For TFIDF, we also use WordPiece tokenization to minimize the number of rare or unknown words.
For consistency, this tokenization uses the same vocabulary as our BERT models do.
FastText is trained to embed whole words directly, so we do not use subword tokenization.

\subsection{Training the Judge}

Here we provide additional implementation details of the various judge models.

\subsubsection{TFIDF}
To limit the number of rare or unknown words, we use subword tokenization via the BERT WordPiece tokenizer.
Using this tokenizer enables us to split sentences in an identical manner as for BERT so that results are comparable.
For a given dataset, we compute inverse document frequencies for subword tokens using the entire corpus.

\subsubsection{BERT}
\paragraph{Architecture and Hyperparameters}
We use the uncased \bertbase pre-trained transformer.
We sweep over BERT fine-tuning hyperparameters, using the following ranges: learning rate $\in \{5\times 10^{-6},1\times 10^{-5},2\times 10^{-5},3\times 10^{-5}\}$ and batch size $\in \{8,12,16,32\}$.

\paragraph{Segment Embeddings}
BERT uses segment embeddings to indicate two distinct, contiguous sequences of input text.
These segments are also separated by a special \texttt{[SEP]} token.
The first segment is $S$, and the second segment is [$Q$; $\Asub{i}$].

\paragraph{Truncating Long Passages}
BERT can only process a maximum of 512 tokens at once.
Thus, we truncate the ends of longer passages; we always include the full question $Q$ and answer $\Asub{i}$, as these are generally important in answering the question.
We include the maximum number of passage tokens such that the entire input (i.e., $(S,Q)$ or $(S,Q,\Asub{i})$) fits within 512 tokens.

\paragraph{Training Procedure}
We train for up to 10 epochs, stopping early if validation accuracy decreases after an epoch once (RACE) or 3 times (DREAM).
For DREAM, we also decay the learning rate by $\frac{2}{3}$ whenever validation accuracy does not decrease after an epoch.

\subsection{Training Evidence Agents}
We use the \bertbase architecture for all learned evidence agents.
The training details are the same as for the BERT judge, with the exceptions listed below.
Agents make sentence-level predictions via end-of-sentence token positions.

\paragraph{Hyperparameters}
Training learned agents on RACE is expensive, due to the dataset size and number of answer options to make predictions for.
Thus, for these agents only (not DREAM agents), we sweep over a limited range that works well: learning rate $\in \{5\times 10^{-6},1\times 10^{-5},2\times 10^{-5}\}$ and batch size $\in \{12\}$.

\paragraph{Training Procedure}
We use early stopping based on validation loss instead of answering accuracy, since evidence agents do not predict the correct answer.

\section{Human Evaluation Details}
\label{sec:Human Evaluation Details}
\begin{figure*}[t]
    \centering
    \includegraphics[width=\textwidth]{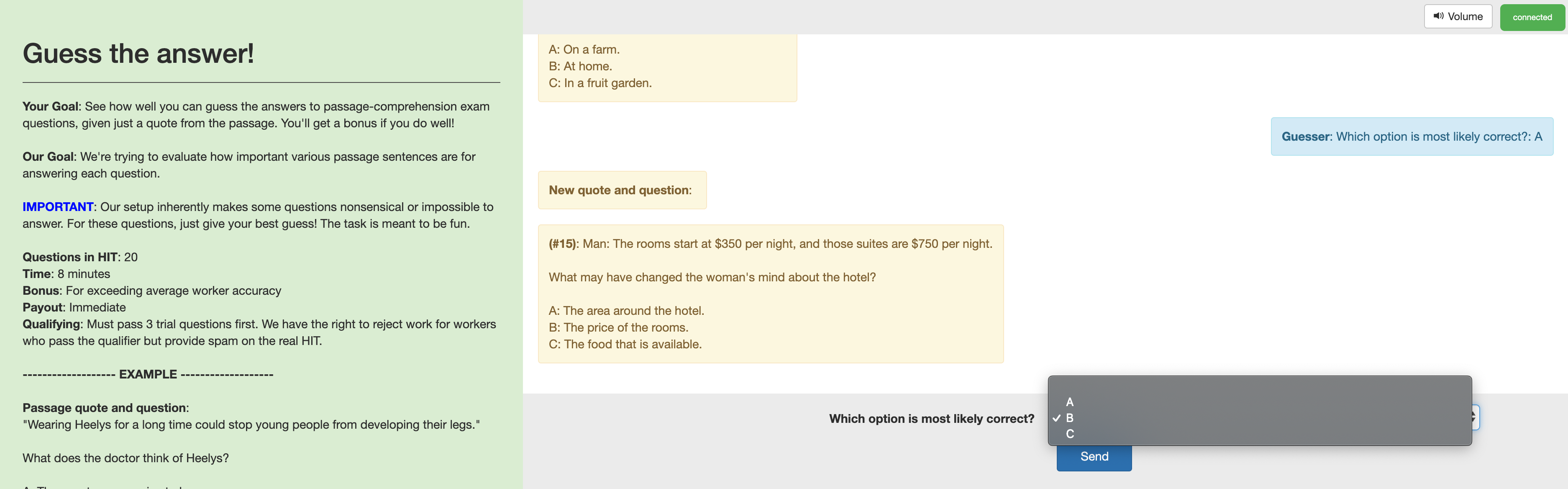}
    \caption{Interface for humans to answer questions based on one agent-selected passage sentence only.
    In this example from DREAM, a learned agent supports the correct answer (B).}
    \label{fig:human_on_quote}
\end{figure*}

\begin{figure*}[t]
    \centering
    \includegraphics[width=\textwidth]{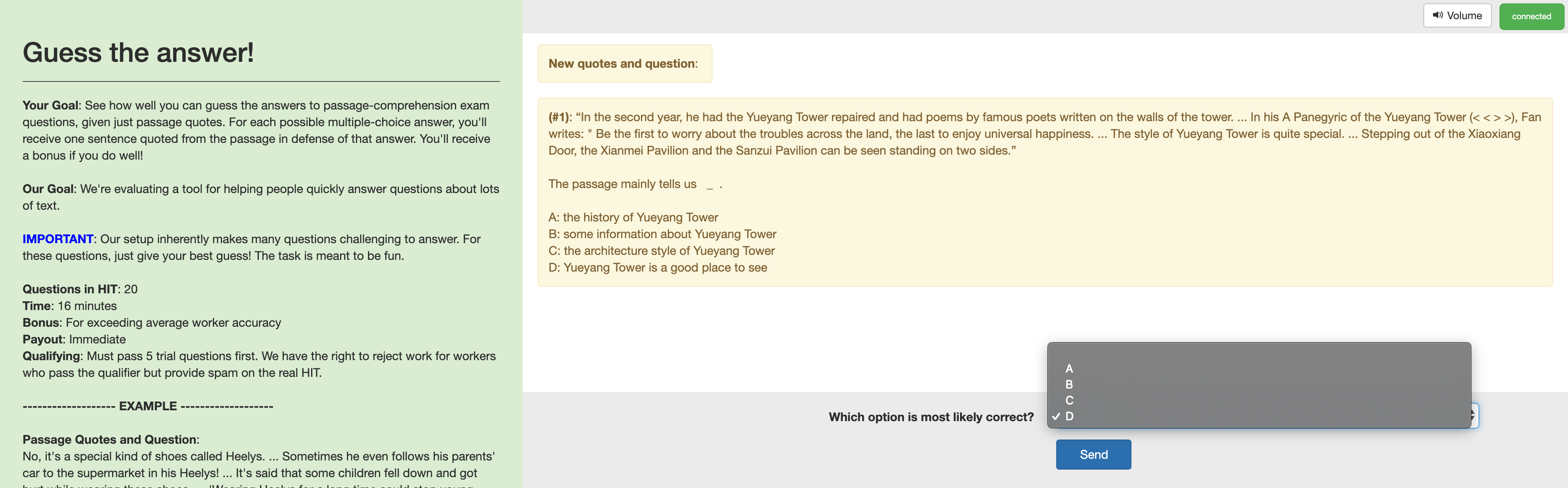}
    \caption{Interface for humans to answer questions based on agent-selected passage sentences only.
    Each answer's evidence agent selects one sentence.
    These sentences are combined and shown to the human, in the order they appear in the passage.
    In this example from RACE, the agents are search-based, and the correct answer is B.}
    \label{fig:human_on_quotes}
\end{figure*}

For all human evaluations, we filter out workers who perform poorly on a few representative examples of the evaluation task.
We pay workers on average \$15.48 per hour according to TurkerView (\url{https://turkerview.com}).
We require workers to be from predominantly English-speaking countries: Australia, Canada, Great Britain, New Zealand, or the U.S.
We do not use results from workers who complete the evaluation significantly faster than other workers (i.e., less than a few seconds per question).
To incentivize workers, we also offer a bonus for answering questions more accurately than the average worker.
Figures~\ref{fig:human_on_quote} and~\ref{fig:human_on_quotes} show two examples of our evaluation setup.

\section{Human Evaluation of Agent Evidence by Question Category}
\label{sec:Human Evaluation of Agent Evidence by Question Category}
\begin{table*}[t]
  \centering
  \resizebox{\textwidth}{!}{
    \begin{tabular} {ll|cccccccc|ccccc}
      \multicolumn{2}{c}{} & \multicolumn{13}{c}{\textbf{\textit{How Often Human Selects Agent's Answer (\%)}}} \\
      \multicolumn{2}{c}{} & \multicolumn{8}{|c}{\textbf{\textit{RACE}}} & \multicolumn{5}{|c}{\textbf{\textit{DREAM}}} \\
      &  & \multicolumn{1}{|c}{} & \multicolumn{2}{c}{\textit{School Level}} & \multicolumn{5}{c}{\textit{Question Type}} & \multicolumn{1}{|c}{} & \multicolumn{4}{c}{\textit{Question Type}} \\
                            & \textbf{Evidence Sentence}     & Overall       & Middle        & High          & Word          & Para-         & Single Sent.  & Multi-Sent.   & Ambi-         & Overall       & Common   & Logic         & Word-Match/      & Summary     \\
                            & \textbf{Selection Method}      &               &               &               & Match         & phrase        & Reasoning     & Reasoning     & guous         &               & Sense    &               & Paraphrase       &             \\
      \midrule
      \textbf{Baselines}  & No Sentence             & 25.0          & 25.0          & 25.0          & 25.0          & 25.0          & 25.0          & 25.0          & 25.0          & 33.3          & 33.3          & 33.3          & 33.3          & 33.3          \\
                            & Human Selection         & 38.1          & 46.4          & 39.5          & 44.6          & 41.3          & 41.7          & 41.7          & 38.5          & 50.7          & 50.0          & 50.6          & 48.2          & 52.1          \\
      \midrule
      \textbf{Search Agents} & TFIDF$(S,[Q;\Asub{i}])$ & 33.5          & 36.5          & 32.2          & 35.0          & 36.1          & 31.8          & 34.2          & 32.7          & 41.7          & 37.2          & 42.4          & 37.1          & 41.8         \\
      \hspace{.25cm} \textbf{querying...}& TFIDF$(S,\Asub{i})$     & 38.0          & 41.8          & 36.4          & 44.8          & 39.9          & 38.4          & 35.2          & 31.1          & 43.4          & 40.0          & 42.7          & 46.4          & 42.7          \\
                            & fastText$(S,\Asub{i})$  & 37.1          & 40.3          & 35.7          & 38.2          & 37.9          & 38.1          & 36.2          & 34.4          & 41.5          & 41.0          & 42.2          & 37.0          & 40.7          \\
                            & \bertbase               & 38.4          & 40.4          & 37.5          & 44.5          & 36.7          & 39.2          & 37.2          & 39.4          & 50.5 & 48.2 & \textbf{50.6} & 52.1 & 50.2 \\
                            & \bertlarge              & 40.1          & 44.5          & 38.3          & 41.3          & 38.8          & 39.9          & \textbf{42.0} & 39.0          & \textbf{52.3} & \textbf{49.8} & 50.3  & \textbf{59.3} & \textbf{54.5} \\
      \textbf{Learned Agents}:  & Search                  & 40.0          & 42.0          & 39.2          & 43.7          & 41.8          & 39.3          & 41.2          & 38.1          & 49.1          & 44.6          & 49.9          & 47.9          & 45.9          \\
      \hspace{.25cm} \textbf{predicting...}& $p(i)$                  & \textbf{42.0} & 44.3          & \textbf{41.0} & \textbf{47.0} & \textbf{43.6} & \textbf{42.3} & 41.9 & 34.3          & 50.0          & 47.6          & 50.1          & 47.3          & 49.6          \\
                            & $\Delta p(i)$           & 41.1          & \textbf{44.9} & 39.5          & 43.7          & 41.4          & 41.0          & 41.9         & \textbf{39.6} & 48.2          & 45.5          & 47.1          & 55.5          & 47.2          \\
    \end{tabular}
  }
  \caption{
    \textit{Human evaluations}:
    \textbf{Search Agents} select evidence by querying the specified judge model, and \textbf{Learned Agents} predict the strongest evidence w.r.t. a judge model (\bertbasenospace); humans then answer the question using the selected evidence sentence (without the full passage).
  }
  \label{tab:human_on_quote_break_down}
\end{table*}

We show a detailed breakdown of results from \S\ref{ssec:Human Evaluation of Evidence}, where humans answer questions using an agent-chosen sentence.
Table~\ref{tab:human_on_quote_break_down} shows how often humans select the agent-supported answer, broken down by question type.
Models that perform better generally do so across all categories.
However, methods incorporating neural methods generally achieve larger gains over word-based methods on multi-sentence reasoning questions on RACE.

\section{Analysis}
\label{sec:Analysis}
\begin{figure}[t]
    \centering
    \includegraphics[width=\columnwidth]{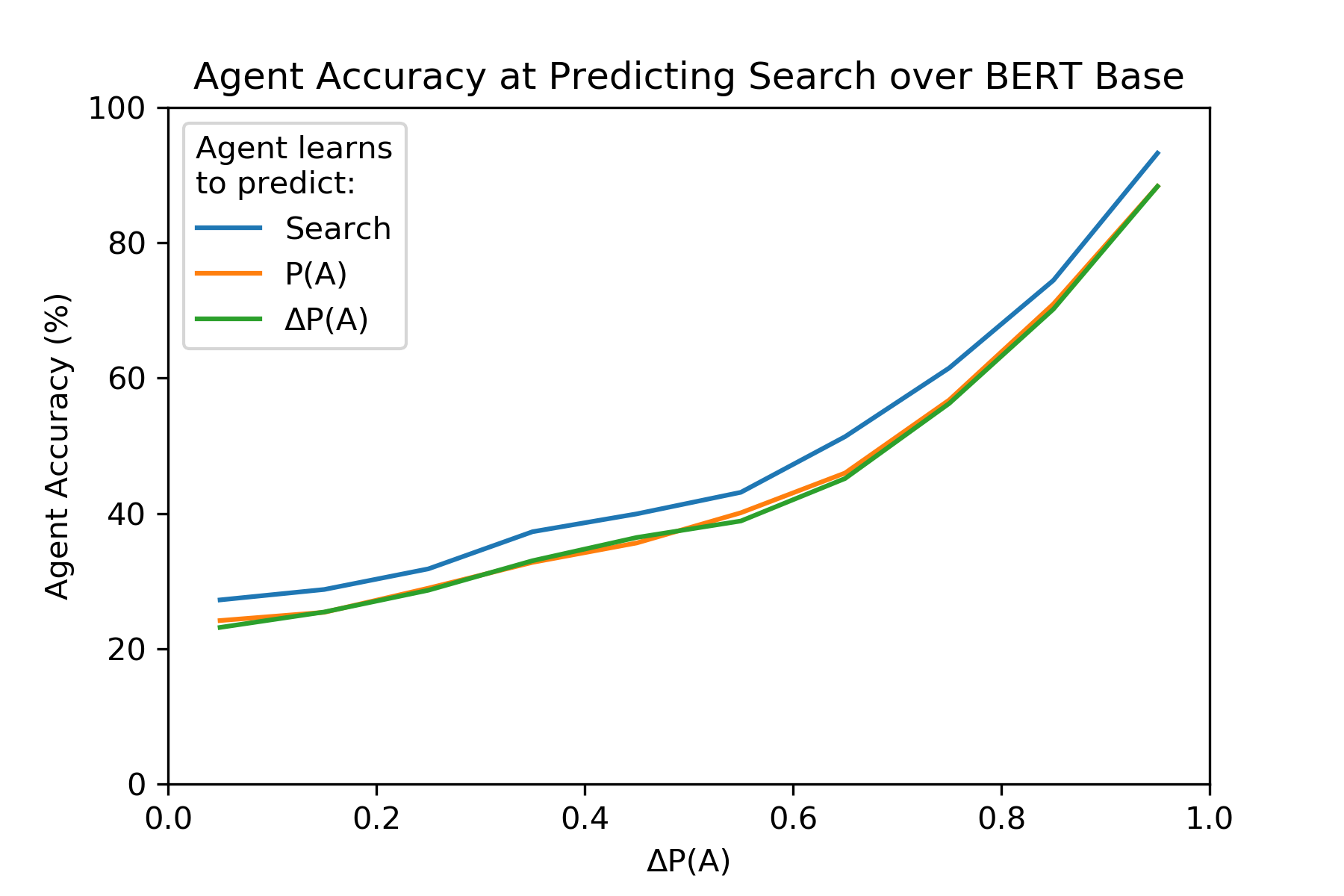}
    \caption{
    Learned agent validation accuracy at predicting the top sentence chosen by search over the judge (\bertbase on RACE).
    The stronger evidence a judge model finds a sentence to be, the easier it is to predict as the being an answer's strongest evidence sentence in the passage.
    This effect holds regardless of the agent's particular training objective.
    }
    \label{fig:accuracy_at_predicting_search}
\end{figure}
\begin{figure}[t]
    \centering
    \includegraphics[width=\columnwidth]{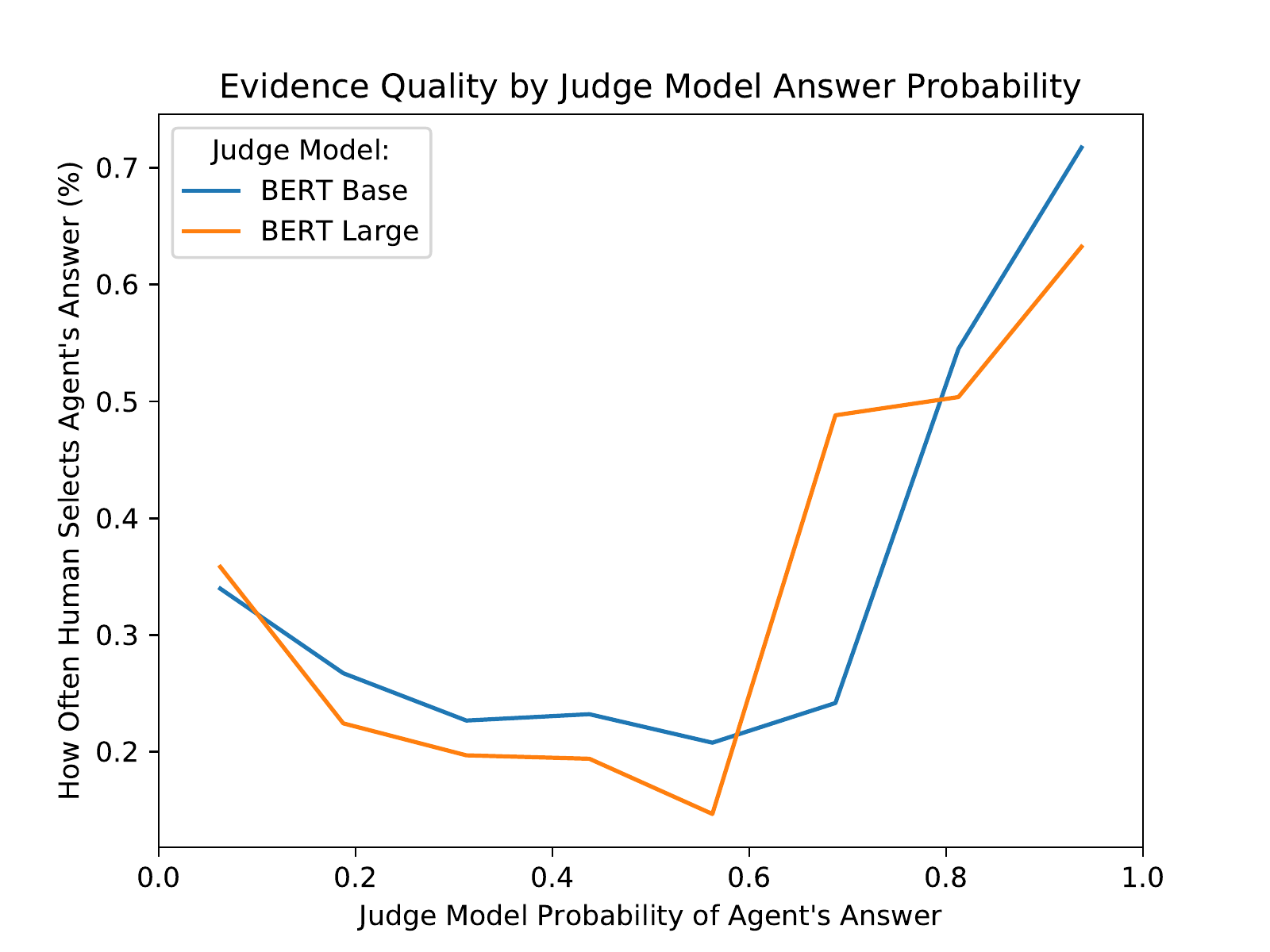}
    \caption{
    We find the passage sentence that would best support an answer to a particular judge model (i.e., using a search agent).
    We plot the judge's probability of the target answer given that sentence against how often humans also select that target answer given that same sentence.
    Humans tend to find a sentence to be strong evidence for an answer when the judge model finds it to be strong evidence.
    }
    \label{fig:evidence_quality_by_judge_model}
\end{figure}

\paragraph{Highly convincing evidence is easiest to predict}
Figure~\ref{fig:accuracy_at_predicting_search} plots the accuracy of a search-predicting evidence agent at predicting the search-chosen sentence, based on the magnitude of that sentence's effect on the judge's probability of the target answer.
Search-predicting agents more easily predict search's sentence the greater the effect that sentence has on the judge's confidence.

\paragraph{Strong evidence to a model tends to be strong evidence to humans} as shown in Figure~\ref{fig:evidence_quality_by_judge_model}.
Combined with the previous result, we can see that learned agents are more accurate at predicting sentences that humans find to be strong evidence.

\section{Model Evaluation of Evidence on DREAM}
\label{sec:Model Evaluation of Evidence on DREAM}
\begin{figure}[t]
    \centering
    \includegraphics[width=\columnwidth]{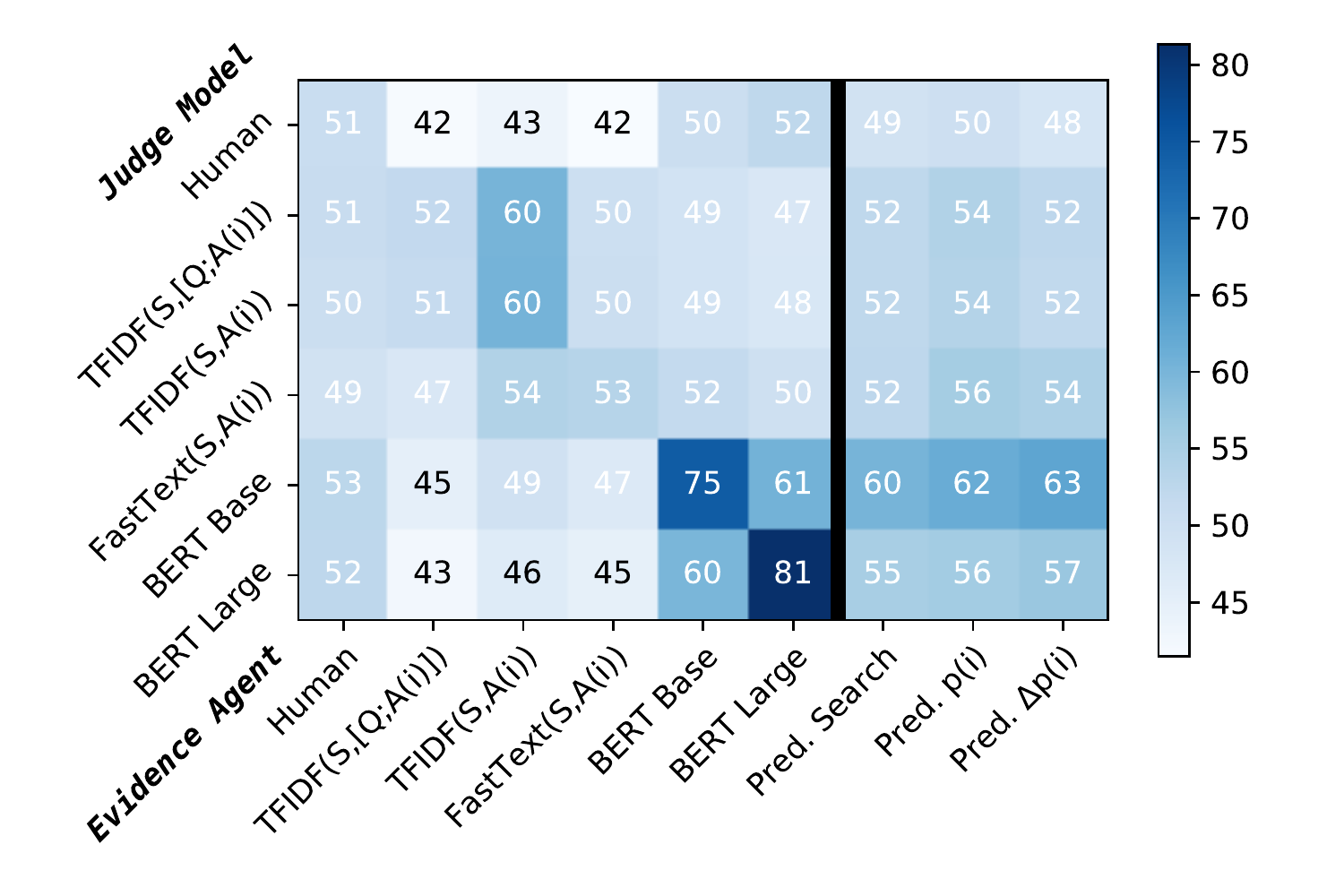}
    \caption{
    On DREAM, how often each judge selects an agent's answer when given a single agent-chosen sentence.
    The black line divides learned agents (right) and search agents (left), with human evidence selection in the leftmost column.
    All agents find evidence that convinces judge models more often than a no-evidence baseline (33\%).
    Learned agents predicting $p(i)$ or $\Delta p(i)$ find the most broadly convincing evidence.
    }
    \label{fig:persuading_models_dream}
\end{figure}

Figure~\ref{fig:persuading_models_dream} shows how convincing various judge models find each evidence agent.
Our findings on DREAM are similar to those from RACE in \S\ref{ssec:Model Evaluation of Evidence}.






\end{document}